\documentclass{article}

\usepackage[nonatbib,final]{neurips_2021}

\usepackage[utf8]{inputenc} 
\usepackage[T1]{fontenc}    
\usepackage{hyperref}       
\usepackage{url}            
\usepackage{booktabs}       
\usepackage{amsfonts}       
\usepackage{nicefrac}       
\usepackage{microtype}      
\usepackage{xcolor}         

\usepackage{graphicx} 
\usepackage{amsmath}
\usepackage{caption,subcaption}
\usepackage{overpic}
\usepackage{physics}
\usepackage{color}
\usepackage{soul}
\usepackage{wrapfig}
\usepackage[numbers,sort&compress]{natbib}
\usepackage{cleveref}
\usepackage[final]{pdfpages}

\usepackage[numbers,sort&compress]{natbib}

\title{Fast Training of Neural Lumigraph Representations using Meta Learning}

\author{%
	Alexander W. Bergman\\
	Stanford University\\
	\small{\texttt{awb@stanford.edu}}
	\And
	Petr Kellnhofer\\
	Stanford University\\
	\small{\texttt{pkellnho@stanford.edu}}
	\And
	Gordon Wetzstein\\
	Stanford University\\
	\small{\texttt{gordon.wetzstein@stanford.edu}}
}

\begin{document}

\maketitle

\vbox{%
	\vskip -0.15in
	\hsize\textwidth
	\linewidth\hsize
	\centering
	\normalsize
	\tt\href{http://www.computationalimaging.org/publications/metanlr/}{computationalimaging.org/publications/metanlr/}
	\vskip 0.28in
}

\newcommand{\R}{\mathbb{R}}
\newcommand{\siren}{{\sc{siren}}}

\newcommand{\loss}{\mathcal{L}}
\newcommand{\bce}{\textsc{bce}}

\newcommand{\image}{I}
\newcommand{\features}{F}
\newcommand{\feature}{f}
\newcommand{\mask}{M}

\newcommand{\extrinsicr}{R}
\newcommand{\extrinsict}{t}
\newcommand{\intrinsic}{C}
\newcommand{\cameraparams}{C}

\newcommand{\shape}{\Phi}
\newcommand{\encoder}{E}
\newcommand{\decoder}{D}
\newcommand{\blending}{\Gamma}
\newcommand{\blendingmlp}{\gamma}
\newcommand{\renderer}{\mathcal{R}}
\newcommand{\warping}{\mathcal{P}}
\newcommand{\weights}{L}

\newcommand{\params}{\Theta}
\newcommand{\paramsshape}{\theta}
\newcommand{\paramsE}{\xi}
\newcommand{\paramsD}{\psi}
\newcommand{\paramsblend}{\zeta}

\newcommand{\viewdir}{v}

\newcommand{\io}{\mathcal{O}}
\newcommand{\red}[1]{\textcolor{red}{#1}}

\newcommand{\ours}{MetaNLR++}
\newcommand{\oursnm}{NLR++}

\definecolor{darkgreen}{rgb}{0.0,0.5,0}
\newcommand{\petr}[1]{\textcolor{darkgreen}{Petr: #1}}
\newcommand{\unsure}[1]{{\sethlcolor{yellow}\hl{#1}}}
\newcommand{\alex}[1]{\textcolor{red}{Alex: #1}}

\begin{abstract}
	Novel view synthesis is a long-standing problem in machine learning and computer vision.
	Significant progress has recently been made in developing neural scene representations and rendering techniques that synthesize photorealistic images from arbitrary views.
	These representations, however, are extremely slow to train and often also slow to render.
	Inspired by neural variants of image-based rendering, we develop a new neural rendering approach with the goal of quickly learning a high-quality representation which can also be rendered in real-time.
	Our approach, \ours{}, accomplishes this by using a unique combination of a neural shape representation and 2D CNN-based image feature extraction, aggregation, and re-projection.
	To push representation convergence times down to minutes, we leverage meta learning to learn neural shape and image feature priors which accelerate training.
	The optimized shape and image features can then be extracted using traditional graphics techniques and rendered in real time.
	We show that \ours{} achieves similar or better novel view synthesis results in a fraction of the time that competing methods require.
	
\end{abstract}
	
\section{Introduction}
\label{sec:intro}
Learning 3D scene representations from partial observations captured by a sparse set of 2D images is a fundamental problem in machine learning, computer vision, and computer graphics. Such a representation can be used to reason about the scene or to render novel views. Indeed, the latter application has recently received a lot of attention (e.g.,~\cite{tewari2020state}). For this problem setting, the key questions are: (1) How do we parameterize the scene, and (2) how do we infer the parameters from our observations efficiently? With our work, we offer new solutions to answer these questions.

Several classes of scene representation learning approaches have recently been proposed. One popular approach consists of coordinate-based neural networks combined with volume rendering, like NeRF~\cite{Mildenhall:2019}. Although these representations offer photorealistic quality for synthesized images, they are slow to train and render. Coordinate-based networks that implicitly model surfaces combined with sphere tracing--based rendering are another popular approach~\cite{Niemeyer2020CVPR,yariv2020multiview,kellnhofer2021neural,oechsle2021unisurf}. One benefit of an implict surface is that, once trained, it can be extracted and rendered in real time~\cite{kellnhofer2021neural}. However, training these representations is equally slow as training volumetric representations. Finally, approaches that use a proxy geometry with on-surface feature aggregation are also fast to render~\cite{riegler2020free,riegler2020stable}, but the quality and runtime of these methods is limited by the traditional 3D computer vision algorithms that pre-compute the proxy shape, such as structure from motion (SfM) or multiview stereo (MVS).

Here, we develop a new framework for neural scene representation and rendering with the goal of enabling both fast training and rendering times. To optimize the training time of our framework, we do not learn a representation network for the view-dependent radiance, as other neural volume or surface methods do, but directly aggregate the features extracted from the source views on the surface of our learned proxy shape.
To cut down on pre-processing times required by SfM and MVS, we optimize a coordinate-based network representing the proxy shape end-to-end with our CNN-based feature encoder and decoder, and learned aggregation function. A key contribution of our work is to combine this unique surface-based neural rendering framework with meta learning, which enables us to learn efficient initializations for all of the trainable parts of our framework and further minimize training time.
Because our representation directly parameterizes an implicit surface, it can be extracted and rendered in real time. This  representation thus incorporates both the fast training benefits from generalizing over shape representations and image features, and the fast rendering capabilities of implicit surface-based methods.
We demonstrate training of high-quality neural scene representations in minutes or tens of minutes, rather than hours or days, which can then be rendered at real-time framerates.

\section{Related Work}
\label{sec:related_work}
\paragraph{Image-based rendering (IBR).}

Classic IBR approaches synthesize novel views of a scene by blending the pixel values of a set of 2D images~\cite{chen1993view,levoy1996light,gortler1996lumigraph,debevec1996modeling,shade1998layered,buehler2001unstructured,shum2000review,chaurasia2013depth,penner2017soft,eslami2018neural}. Recent IBR techniques leverage neural networks to learn the required blending weights~\cite{HPPFDB18,choi2019extreme,chen2019monocular,meshry2019neural,thies2019deferred,bemana2020x}. These neural IBR methods either use proxy geometry, for example obtained by SfM or MVS~\cite{schoenberger2016sfm,schoenberger2016mvs} or depth estimation~\cite{wiles2020synsin}, together with on-surface feature aggregation~\cite{riegler2020free,riegler2020stable}, or use learned pixel aggregation functions~\cite{yu2020pixelnerf,wang2021ibrnet} for geometry-free image-based view synthesis. 
Our approach is closely related to the geometry-assisted and feature-interpolating view synthesis techniques. Many existing approaches, however, require the proxy geometry to be estimated as a pre-processing step, which can take minutes to hours for a single scene, preventing fast processing of novel views. Other approaches which use monocular depth estimation to re-project and aggregate features such as SynSin~\cite{wiles2020synsin}, are applied only to single input images, and cannot easily be scaled to multi-view data due to depth estimation inconsistencies between images of the scene. Instead, our approach estimates a coordinate-based neural shape representation from the input images. The shape representation is optimized end-to-end with the image feature extraction, aggregation, and decoding and is accelerated using meta learning.

\paragraph{Neural scene representations and rendering.}

Emerging neural rendering techniques~\cite{tewari2020state} use explicit, implicit, or hybrid scene representations. Explicit representations, for example those using proxy geometry (see above), object-specific shape templates~\cite{kanazawa2018learning}, multi-plane~\cite{Zhou:2018,Mildenhall:2019,flynn2019deepview,tucker2020single,Wizadwongsa2021NeX} or multi-sphere~\cite{Broxton:2020,Attal:2020:ECCV} images, or volumes~\cite{sitzmann2019deepvoxels,Lombardi:2019}, are fast to evaluate but memory inefficient. Implicit representations, or coordinate-based networks, can be more memory efficient but are typically slow to evaluate~\cite{park2019deepsdf,mescheder2019occupancy,genova2019learning,genova2019deep,chen2019learning,michalkiewicz2019implicit,atzmon2019sal,saito2019pifu,sitzmann2019srns,Oechsle2019ICCV,gropp2020implicit,yariv2020multiview,davies2020overfit,sitzmann2020siren,Niemeyer2020CVPR,liu2020neural,jiang2020sdfdiff,liu2020dist,kohli2020inferring,kellnhofer2021neural}. Hybrid representations make a trade-off between computational and memory efficiency~\cite{peng2020convolutional,jiang2020local,chabra2020deep,martel2021acorn}. 
Among these, NeRF~\cite{mildenhall2020nerf} and related methods (e.g.,~\cite{martinbrualla2020nerfw,niemeyer2020giraffe,pumarola2020d,srinivasan2020nerv,zhang2020nerf,neff2021donerf}) use coordinate-based network representations and volume rendering, which requires many samples to be processed along a ray, each requiring a full forward pass through the network. Although recent work has proposed enhanced data structures~\cite{yu2021plenoctrees,hedman2021snerg,garbin2021fastnerf}, network factorizations~\cite{Reiser2021ICCV}, pruning~\cite{liu2020neural}, importance sampling~\cite{neff2021donerf}, fast integration~\cite{lindell2020autoint}, and other strategies to accelerate the rendering speed, training times of all of these methods are extremely slow, on the order of hours or days for a single scene. DVR~\cite{Niemeyer2020CVPR}, IDR~\cite{yariv2020multiview}, NLR~\cite{kellnhofer2021neural} and UNISURF~\cite{oechsle2021unisurf} on the other hand, leverage implicitly defined surface representations, which are faster to render than volumes but are equally slow to train. While concurrent work has improved the applicability of these methods by addressing limitations of surface-based methods in general, such as removing the object mask requirement~\cite{oechsle2021unisurf, yariv2021volume, wang2021neus}, these approaches are still slow to train as they rely on hybrid volumetric and surface formulations to bootstrap the neural surface training. Our approach builds on generalization approaches for neural scene representations to accelerate the training time of 2D-supervised neural surface representations, and thus can be applied alongside advanced training strategies for neural surfaces and make these representations even more applicable for practical view synthesis.

\paragraph{Generalization with neural scene representations.}

Being able to generalize across neural representations of different scenes is crucial for learning priors and for 3D generative-adversarial networks with coordinate-based network backbones~\cite{hologan,nguyen2020blockgan,graf,chanmonteiro2020piGAN,niemeyer2021campari}. A variety of different generalization approaches have been explored for neural scene representations, including conditioning by concatenation~\cite{mescheder2019occupancy,park2019deepsdf}, hypernetworks~\cite{sitzmann2019srns}, modulation or feature-wise linear transforms~\cite{chanmonteiro2020piGAN,chen2020learning,mehta2021modulated}, and meta learning~\cite{sitzmann2019metasdf,tancik2020learned}. Inspired by these works, we propose a meta-learning strategy that allows us to quickly learn a high-quality neural shape representation for a given set of multi-view images. As opposed to the 3D point cloud supervision proposed by MetaSDF~\cite{sitzmann2019metasdf}, we meta-learn signed distance functions (SDFs) using 2D images, and as opposed to meta-learned initializations for NeRF volumes~\cite{tancik2020learned}, we operate with SDFs and features. Our approach is unique in enabling fast, high-quality shape representations to be learned from multi-view images that, once optimized, can be rendered in real time.


\section{Method}
\label{sec:method}
\subsection{Image formation model}
\begin{figure}
	\includegraphics[width=1\textwidth]{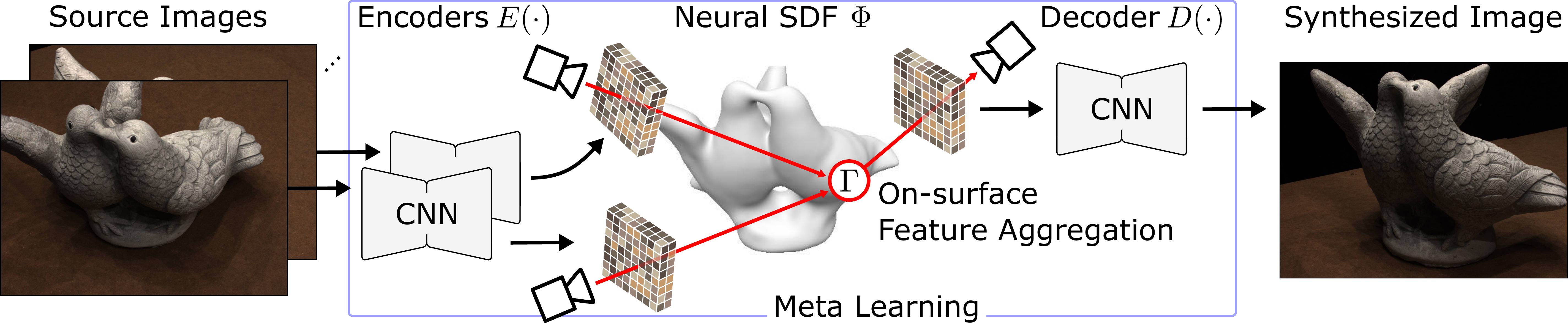} 
	\caption{Overview of \ours{}.}
	\label{fig:methods}
\end{figure}

In this section, we outline the \oursnm{} novel view synthesis image formation model, which is presented in Figure~\ref{fig:methods}. 
\oursnm{} takes as inputs a set of images $\{\image_i\}_{i=1}^N$ each with corresponding binary foreground object masks $\{\mask_i\}_{i=1}^N$ and known camera intrinsic parameters and extrinsic parameters collectively referred to as $\cameraparams_i$. At output, \oursnm{} synthesizes an image $\hat{\image}_t$ of the scene from the target viewpoint $\cameraparams_t$.

Drawing inspiration from classic image-based rendering methods, we define the image formation model as a learned pixel-wise aggregation $\blending_\paramsblend(\cdot)$ of input image features $\{\encoder_\paramsE(\image_i)\}_{i=1}^N$ and their target viewing direction $V_t\in\R^{H\times W\times 3}$ on the surface of the object represented by the neural surface $\shape_\paramsshape$. 
To project the non-occluded, visible input features of $\encoder_\paramsE(\image_i)$ into the target view before aggregation, we define the function $\warping_{i\rightarrow t}(\encoder_\paramsE(\image_i); \cameraparams_i, \cameraparams_t)$.
These neurally aggregated features are then decoded into an image by decoder $\decoder_\paramsD(\cdot)$:
\begin{equation}
	\hat{\image}_t = \decoder_\paramsD(\blending_\paramsblend(\{\warping_{i\rightarrow t}(\encoder_\paramsE(\image_i))\}_{i=1}^N, V_t)).
\end{equation}
The feature encoder $\encoder$ and decoder $\decoder$ are implemented as resolution-preserving convolutional neural network (CNN) architectures~\cite{he2015deep, ronneberger2015unet} with learned parameters $\paramsE, \paramsD$:
\begin{equation}
	\encoder_\paramsE: \R^{H\times W\times 3}\rightarrow \R^{H\times W\times d}, \hspace{1em} \decoder_\paramsD: \R^{H\times W\times d}\rightarrow \R^{H\times W\times 3}
\end{equation}

To aggregate the input image features from $\encoder$ into a target feature map to be decoded by $\decoder$, we use a learned feature aggregation (or blending) function $\blending_\paramsblend$, which operates on the surface of our shape representation $\shape$.
To define the surface of our shape, we choose to use a \siren~\cite{sitzmann2020siren} which represents the signed-distance function (SDF) in 3D space. This encodes the surface of the object as the zero-level set, $L_0$, of the network:
\begin{equation}
	L_0(\shape_\paramsshape) = \{x\in\R^3 | \shape_\paramsshape(x)=0 \}, \hspace{1em} \shape_\paramsshape:\R^3\rightarrow\R.
\end{equation}

The aggregation is performed on surface for each pixel of the target image $\hat{\image_t}$ with camera $\cameraparams_t$. 
To find the point in $L_0(\shape_\paramsshape)$ corresponding to each pixel ray, we perform sphere tracing on the neural SDF model $\renderer(\shape_\paramsshape, \cameraparams_t)$, retaining gradients for the final step of evaluation~\cite{yariv2020multiview, kellnhofer2021neural, liu2020dist, jiang2020sdfdiff}.
These individual rendered surface points are projected into the image plane of each of the $N$ input image views and used to sample interpolated features from the source feature maps for each pixel, which can be arranged into $N$ re-sampled feature maps corresponding to each input image $\{\features_i\}_{i=1}^N\in\R^{N\times H\times W\times d}$. 
To check whether or not a feature is occluded, we use sphere tracing for each pixel from the input views $\{\renderer(\shape_\paramsshape, \cameraparams_i)\}_{i=1}^N$, and ensure that the target view surface position projected into each of these surfaces is at the same depth as the source view surface position. Occluded features are then discarded.
These three steps of sphere tracing, feature sampling, and occlusion checking make up the function  $\warping_{i\rightarrow t}(\encoder_\paramsE(\image_i))$ which outputs each of the re-sampled feature maps $\{\features_i\}_{i=1}^N$.

Once the re-sampled feature maps $\{\features_i\}_{i=1}^N$ have been generated, the aggregation function $\blending_\paramsblend$ aggregates them into a single target feature $\features_{t}\in\R^{H\times W\times d}$ which can be processed by the decoder into an image. This is done by performing a weighted averaging operation on the input features using the relative feature weights $\weights\in\R^{N\times H\times W}$:
\begin{gather}
	\blending_\paramsblend: \R^{N\times H\times W\times d}\times\R^{H\times W\times 3}\rightarrow\R^{H\times W\times d}, \hspace{1em} \nonumber\\
	\{\features_i\}_{i=1}^N, V_t\mapsto\blending_\paramsblend(\{\features_i\}_{i=1}^N, V_t) = \sum_{i=1}^N (\weights_i / \sum_{j=1}^N \weights_j) \circ\features_i = \features_{t},
\end{gather}
where $\circ$ is the Hadamard product between the feature and weight maps, broadcasted over the feature dimension $d$.
The weight map $\weights$ used in the feature aggregation function $\blending_\paramsblend$ is obtained from an MLP $\blendingmlp_\paramsblend$ which is applied pixel-wise to each of the $N$ re-sampled feature maps and each pixels target viewing direction:
\begin{equation}
	 [\{\features_i\}_{i=1}^N, V_t]\mapsto\blendingmlp_\paramsblend([\{\features_i\}_{i=1}^N, V_t]) = \weights. 
\end{equation}
Here, the dependence upon viewing direction allows for the modeling of view-dependent image properties. These pixel-wise operations making up $\blending$ result in a $H\times W\times d$ feature map which can be input into the decoder $\decoder$.

The usage of features instead of pixel values directly allows $\decoder$ some opportunity to inpaint and correct artifacts from imperfect geometry to create a photorealistic novel view, unlike methods which render pixels individually~\cite{yariv2020multiview, kellnhofer2021neural}.
Additionally, the use of the CNN encoder and decoder increases the receptive field of the image loss applied, allowing for more meaningful gradients to be propagated back into $\shape_\paramsshape$.

\subsection{Supervision and training}
Since \oursnm{} is end-to-end differentiable, we can optimize the parameters $\paramsE, \paramsD, \paramsshape, \paramsblend$ end-to-end to reconstruct target views.
For each iteration of training we sample a set of $k \leq N$ images $\{\image_n\}_{n=1}^k$, and designate one of these images to be the ground-truth target image used for supervision $\image_t$, and sample its corresponding binary object mask $\mask_t$ and parameters $\cameraparams_t$.

Using the \oursnm{} image formation model, we generate a synthesized target image $\hat{\image}_t$ from viewpoint $\cameraparams_t$.
The loss evaluated on the synthesized image consists of three terms:
\begin{equation}
	\loss(\{\hat{\image}_t, \hat{\mask}_t\}, \{\image_t, \mask_t\}, \shape_\paramsshape) =
	\loss_{R}(\hat{\image}_t, \{\image_t, \mask_t\}) 
	+ \lambda_1\loss_{M}(\shape_\paramsshape, \mask_t) 
	+ \lambda_2\loss_{E}(\shape_\paramsshape).
\end{equation}
The image reconstruction loss $\loss_{R}$ is computed as a masked L1 loss on rendered images:
\begin{equation}
	\loss_{R}(\hat{\image}_t, \{\image_t, \mask_t\}) = \frac{1}{\sum_p \mask_t[p]} \sum_{p|\mask_t[p]=1} |\image_t[p] - \hat{\image_t}[p]|.
\end{equation}
To quickly bootstrap the neural shape learning from the object masks, we apply a soft mask loss on the rendered image masks~\cite{kellnhofer2021neural, yariv2020multiview}:
\begin{equation}
	\loss_{M}(\shape_\paramsshape, \mask_t) = \frac{1}{\alpha\sum_p \mask_t[p]} \sum_{p|\mask_t[p]=0 \lor \shape_{min}[p]<\tau} \bce(\text{sigmoid}(-\alpha \shape_{min}[p], \mask_t[p])),
\end{equation}
where the notation $\shape_{min}[p]$ denotes the minimum value of the SDF $\shape_\paramsshape$ along the ray traced from pixel $p$, $\tau$ is a threshold for whether the zero level set $L_0(\shape_\paramsshape)$ has been intersected, and $\alpha$ is a softness hyperparameter.
Finally, we regularize the shape representation to model a valid SDF by enforcing the eikonal constraint on randomly sampled points $p_i\in\R^3$ in a unit cube containing the scene:
\begin{equation}
	\loss_{E}(\shape_\paramsshape) = \frac{1}{P} \sum_{i=0}^P ||(||\nabla_p\shape_\paramsshape(p_i)||_2 - 1)||_2^2
\end{equation}

However, to make our training more efficient, we augment the loss supervision schedule and batching strategy for our model.
Specifically, for each sampled batch of $k$ images, instead of computing gradients for a single selected target image, we treat $l<k$ images as target images, and each of them from the other images in the batch. The number of target images $l$ is limited by the GPU memory when computing the loss for each target image.
Since all views must be sphere traced and passed through $\encoder$ for a single target view, this additional batching only adds additional forward passes to $\blending$ and $\decoder$, which are fast to evaluate. 
This batching strategy gives more accurate gradients for our model at each iteration.
Additionally, since $\loss_{M}$ requires us to find a minimum of $\shape_\paramsshape$ along a ray, it requires dense sampling of this network and accounts for most of the compute time of each forward pass.
Thus, while optimizing \oursnm{}, we propose to only enforce shape-related losses $\loss_{M}$ and $\loss_{E}$ for the first $t_1$ iterations, and then every $t_2$ iterations thereafter. 
This allows \oursnm{} to learn a shape approximation in the first $t_1$ iterations, and then further refine it as the feature encoding scheme with $\encoder,\decoder,\blending$ get significantly better. 
This is only possible since, unlike prior Neural Lumigraph work~\cite{kellnhofer2021neural, yariv2020multiview}, the appearance modeling is outsourced to the feature extraction from input images and is more independent from the current shape than a dense appearance representation in 3D space.

\subsection{Generalization using meta learning}
As our goal is to learn scene representations quickly, we use meta learning to learn a prior over feature encoding, decoding, aggregation, and shape representation using datasets of multi-view images.
This prior is realized via the initialization of the networks $\encoder_\paramsE, \decoder_\paramsD, \blending_\paramsblend$ and $\shape_\paramsshape$ which dictates the network convergence properties during gradient descent optimization.
For simplicity of notation and since we are meta-learning the initializations for all networks in \oursnm{}, we define all \oursnm{} parameters as $\params = [\paramsE, \paramsD, \paramsblend, \paramsshape]$.

Let $\params_0$ denote the \oursnm{} parameters at initialization, and $\params_i$ denote the parameters after $i$ iterations of optimization. 
For a fixed number of steps $m$ of optimization, $\params_m$ will depend significantly on the initialization $\params_0$, resulting in possibly significantly different \oursnm{} losses. 
We adopt the notation from ~\cite{tancik2020learned}, and will emphasize the dependence of parameters on initialization by writing $\params_m(\params_0, T)$, where $T$ is the particular scene which we would like to represent.
We aim to optimize the initial weights $\params_0$ that will result in the lowest possible expected loss after $m$ iterations when optimizing \oursnm{} for an unseen object $T$, sampled from a distribution of objects $\mathcal{T}$. This expectation over objects is denoted as $E_{T\sim\mathcal{T}}$, resulting in the meta learning objective of:
\begin{equation}
	\params_0^* = \arg\min_{\params_0}E_{T\sim\mathcal{T}}[\loss(\params_m(\params_0, T))].
\end{equation}
To learn this initialization for $\params_0$, we use the Reptile~\cite{nichol2018firstorder} algorithm, which computes the weight values $\params_m(\params_0, T)$ for a fixed inner loop step size of $m$.
Each optimization of \oursnm{} for $m$ steps for a different task sampled from $T_j\sim\mathcal{T}$ is referred to as an outer loop, and is indexed by $j$.
To avoid having to compute second-order gradients through the \oursnm{} model, Reptile updates the initialization $\params_0$ in the direction of the optimized task weights with the following equation:
\begin{equation}
	\params_0^{j+1} = \params_0^j + \beta(\params_m(\params_0^j, T_j) - \params_0^j),
\end{equation}
where $\beta$ is the meta-learning rate hyperparameter.
We label \oursnm{} with the meta-learning initializations applied \ours{}.

\begin{figure}
	\includegraphics[width=\textwidth]{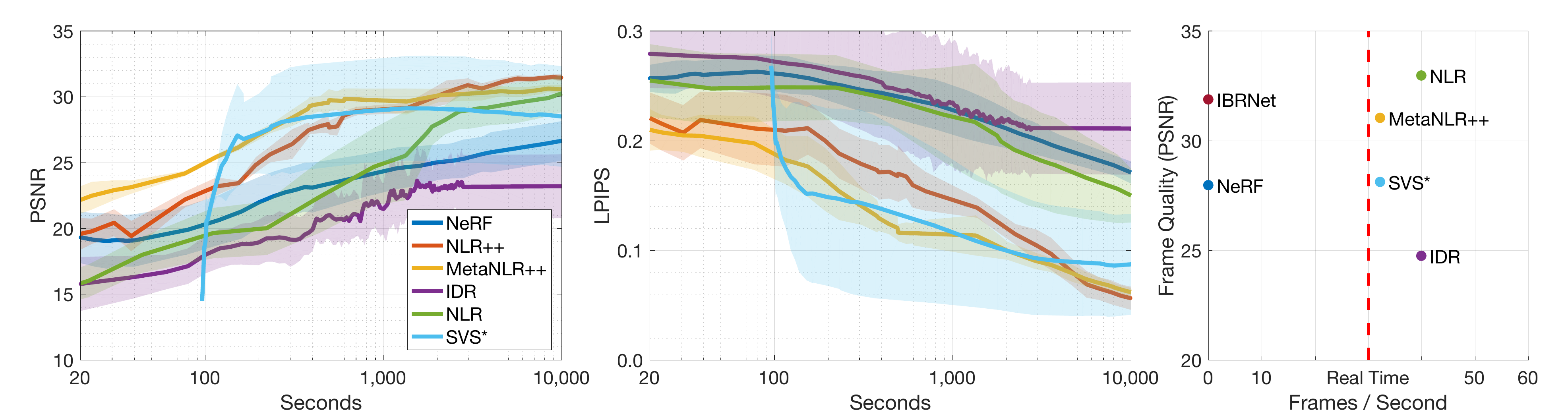} 
	\caption{We demonstrate that at all training-times, \ours{} is comparable to or outperforms all competitive representation learning methods, including both neural volumetric and surface representations in PSNR$\uparrow$ (left) and LPIPS$\downarrow$~\cite{zhang2018unreasonable} (center). We also plot the render time versus converged image quality, showing that \ours{} can generate high-quality frames at real-time rates (right). The shaded area around each line represents the standard deviation of the method across three DTU scenes.}
	\label{fig:trade_off}
\end{figure}

\subsection{Implementation details}
\label{sec:implementation_details}
The source code and pre-trained models are available on our project website, and the full set of implementation details including hyperparameters, training schedules, and architectures are described in our supplement for each of the various datasets evaluated on.
We implement \ours{} in PyTorch and use the Adam~\cite{kingma2014adam} optimizer for all optimization steps of \oursnm{}, including for the inner-loop in meta learning, with a starting learning rate of $1\times10^{-4}$ for $\shape$ and $5\times10^{-4}$ for $\encoder,\decoder,\blending$.
We use $\alpha=50$, $\tau=1\times10^{-3}$, and $\beta=1\times10^{-1}$ as starting hyperparameter values, which are progressively decayed (or increased in the case of $\alpha$) through training (full schedules are described in the supplement). We use shape loss training hyperparameter values of $t_1=50$ and $t_2=7$, and loss weight parameters of $\lambda_1=1\times10^{2}/\alpha$, $\lambda_2=3$.
In the case of \oursnm{}, we initialize $\shape$ as a unit sphere of radius $0.5$ by pre-training our {\siren} to represent this shape.
We train each of our models on a single Nvidia Quadro RTX8000 GPU. We also use an Nvidia Quadro RTX6000 GPU for rendering and training iteration time computation.
In total, we have an internal server system with four Nvidia Quadro RTX8000 GPUs and six Nvidia Quadro RTX6000 GPUs, of which we used a subset of three RTX8000s and one RTX6000.

\section{Experiments}
\label{sec:experiments}

\paragraph{Baselines.}
Our main contribution is the rapid learning of a representation which can be used to render high-quality novel views of a scene in real time.
We demonstrate this by comparison to several state-of-the-art methods. Specifically, we evaluate the volumetric representation of NeRF~\cite{mildenhall2020nerf}, a mesh-based representation similar to SVS~\cite{riegler2020stable}, the neural signed distance function-based representations of IDR~\cite{yariv2020multiview} and NLR~\cite{kellnhofer2021neural}, and the image-based rendering of IBRNet~\cite{wang2021ibrnet}.
For SVS~\cite{riegler2020stable} we use our own simplified implementation and denote it SVS*. Our implementation trains the same $\encoder,\decoder,\blending$ as in \ours{} but we replace the learnable shape by a surface reconstruction from COLMAP~\cite{schoenberger2016sfm, schoenberger2016mvs}.
\begin{figure}
	\includegraphics[width=\textwidth]{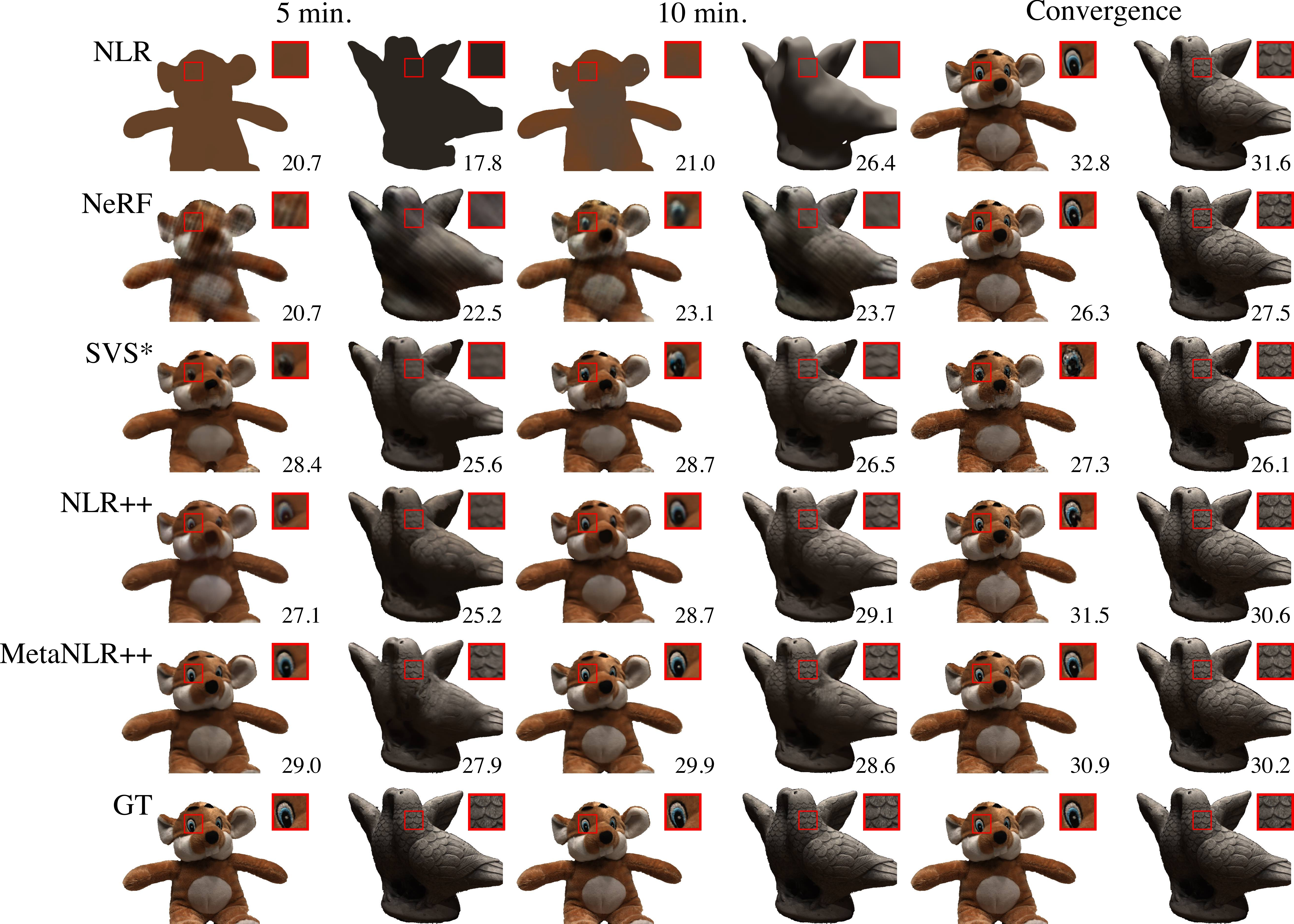} 
	\caption{Novel views synthesized using various methods after a set training time. \ours{} outperforms other surface and volume representation methods, especially for a training time budget on the order of minutes, and does not sacrifice quality of the final converged result.}
	\label{fig:qual_comp}
\end{figure}

\paragraph{Training time vs. quality trade-off.}
For the following comparisons, we use the DTU dataset~\cite{jensen2014large,yariv2020multiview}, which has been made public by its creators, and contains multi-view images of various inanimate objects, none of which are offensive or personally identifiable.
In Figure~\ref{fig:trade_off}, we plot the average PSNR and LPIPS score of three held-out test views on three test DTU scenes as a function of training time as measured on a Nvidia Quadro RTX6000.
Each of these representations are trained using only seven ground-truth views from the DTU scene.
The meta-learned initializations $\params_0^*$ are optimized using complete view sets from another 15 DTU scenes, distinct from the testing scenes. 
In these plots, we see that \ours{} maintains high reconstruction quality throughout the training process which results in predictable quality progression for time-constrained applications.
Beyond PSNR we showcase results of the perceptual LPIPS metric~\cite{zhang2018unreasonable} as we observe that PSNR is not robust to small inaccuracies in the object masks and prefers low-frequency images.
This trade-off is exemplified in Table~\ref{tab:quantitative}, where we show that \ours{} is able to reach the 25dB and 30dB PSNR milestones faster than any other learned scene representation.
The difference is particularly large for the volumetric method NeRF that aggregates many radiance samples along each rendered ray, and the implicit surface-based method NLR which must simultaneously optimize a neural surface and neural representation of color densely in 3D. In the case of NLR, this is because every time the shape representation is updated, the color representation must also update to correctly reflect the appearance on the modified surface of the shape, leading to a slow training time.
\begin{wraptable}[18]{r}{0.7\textwidth}
	\vspace{-1em}
	\caption{Training time to reach specified PSNR level. The best times and PSNR values are bolded (second best is underlined) for methods which can render at real-time framerates. The training times are included for NeRF and IBRNet as well, which are incapable of fast rendering. IBRNet is pre-trained on multi-view data, and thus needs no further training to reach 25dB PSNR.}
	\label{tab:quantitative}
	\centering
	\begin{tabular}{lccc}
		\small
		& 25dB PSNR & 30dB PSNR & Maximum PSNR \\
		\midrule
		IDR & - & - & 24.73dB \\ 
		NLR & 14.7 min. & 191.4 min. & \textbf{32.95dB} \\
		MetaNLR & \underline{2.1 min.} & 176.8 min. & \underline{31.71dB} \\
		SVS* & \underline{2.1 min.} & - & 28.19dB \\ 
		\oursnm{} & 3.2 min. & \underline{37.4 min.} & 31.02dB \\
		\ours{}  & \textbf{1.9 min.} & \textbf{22.5 min.} & 30.57dB \\
		\midrule
		NeRF & 33.3 min. & - & 27.95dB \\ 
		IBRNet & 0 sec. & 23.1 sec. & 31.86dB \\
		\bottomrule
	\end{tabular}
\end{wraptable}
While IBRNet~\cite{wang2021ibrnet} is able to generate high-quality novel views very quickly, it cannot be turned into a pre-computed mesh-based or volume representation and requires input images for each rendered frame, and is more aptly considered a feed-forward image-blending method instead of a neural scene representation.
SVS* is initially offset by the runtime of COLMAP. 
Afterwards, it quickly fits $\encoder,\decoder,\blending$, but its maximum performance is limited by the initial geometry. Consequently, $\decoder$ must learn to inpaint the resulting holes which results in over-fitting to training views. This is notable in Figures~\ref{fig:trade_off}, \ref{fig:qual_comp} and Table~\ref{tab:quantitative}, where the quality of novel views quickly saturates or even degrades.
Additionally, the meshing step time scales with the number of input views, and it can take up to 2 hours for scenes with dense view coverage as reported in \cite{riegler2020stable}.
MetaNLR is provided as a baseline which applies our meta learning method to the NLR formulation. This shows that simply applying meta learning to NLR does not lead to nearly as fast of training to high quality when compared to \ours{}, and therefore the improvements in scene parameterization in NLR++ and the generalization via meta learning are necessary components which jointly contribute to achieving the desired goal of both fast training and rendering. Qualitative comparisons are shown in the supplementary document.
We emphasize that learning a representation from only seven views is a difficult task.
NeRF in particular has difficulty avoiding over-fitting to training views in this scenario.
We show qualitative results in Figure~\ref{fig:qual_comp}, which highlight that \ours{} performs progressively higher quality view synthesis using the learned features and geometry as the training advances.

\paragraph{Training to convergence.}
In Table~\ref{tab:quantitative} and Figure~\ref{fig:qual_comp}, we show that \ours{} does not sacrifice on final converged quality for the sake of speed. Given unconstrained learning time, \ours{} is still able to produce images which are competitive with converged results of state-of-the-art representations.
As noted in Table~\ref{tab:quantitative}, IBRNet is able to also perform high quality novel view synthesis, and can render images with $29.20$dB without fine-tuning. However, since the rendering time is based off of neural volume rendering, it cannot generate frames at nearly a real-time rate. 
Additional comparisons are provided in the supplementary document.
A quantitative evaluation of image quality and runtime at rendering time is shown in Figure~\ref{fig:trade_off}, where we see that our surface-based method is able to render in real time, unlike neural volume rendering methods.

\paragraph{Real-time rendering.}
Because \ours{} extracts the appearance directly from the source images and transforms them to the novel view using a compact shape model, we can greatly accelerate the rendering of our trained representation by pre-computing these components.
First, we use the marching cubes algorithm~\cite{lorensen1987marchingcubes} to extract a mesh as an iso-surface of the SDF. Second, we store the features computed from the set of input images by our encoder $\encoder$ as well as their associated depth maps.
At render time, we simply need to aggregate these features based on our target viewing direction, and evaluate the decoder on the aggregated features for each frame. For the DTU images and network architecture, the decoder takes $8.7$ms, and the aggregation takes $22.3$ms, so we are able to render frames at real-time rates as shown in Figure~\ref{fig:trade_off}.

\paragraph{Meta-learning ablation.}
We investigate the efficacy of meta learning in speeding up learning of individual components of our representation.
Specifically, we ablate the contribution of meta learning on the shape representation parameters ($\paramsshape_0$), and feature encoding network parameters ($\paramsE_0, \paramsD_0, \paramsblend_0$). 
\begin{wraptable}[10]{r}{0.63\textwidth}
	\vspace{-.3cm}
	\caption{Meta-learning ablation study. We report average PSNR of synthesized views after 10 minutes for each method.}
	\label{tab:ablation}
	\centering
	\vspace{-0.2cm}
	\begin{tabular}{lccc}
		\small
		& \multicolumn{3}{c}{Meta init. applied to:} \\
		& No net. & $\shape$ only & All net. \\
		\midrule
		RGB pixels \& fixed $\blending$ & 27.85dB & 28.13dB & 28.13dB  \\ 
		RGB pixels \& learned $\blending$ & 26.82dB & 26.90dB & 26.91dB \\
		CNN $\encoder$/$\decoder$ \& fixed $\blending$ & 28.62dB & 29.16dB & 29.88dB \\
		CNN $\encoder$/$\decoder$ \& learned $\blending$ & 27.32dB & 28.48dB & \textbf{29.91dB} \\
		\bottomrule
	\end{tabular}
\end{wraptable}
Additionally, we also compare \ours{} to a variant with the encoder and decoder replaced by a direct extraction of image RGB pixel values and with and without a learned aggregation function. Here we clarify that replacing the encoder and decoder with pixel values is still a variant of NLR++, where we have simply replaced the features with pixel values directly, and does not model a dense, continuous color function as in NLR.
The results of this ablation study are shown in Table~\ref{tab:ablation}. 

We see that meta-learning of each component contributes to fast learning of a high-quality representation (last row).
The learned $\blending$ improves the performance for the learned $\encoder$/$\decoder$ but it decreases the performance for the directly extracted RGB values.
This implies that the classical unstructured lumigraph blending~\cite{buehler2001unstructured} works well for direct pixel values, but the additional flexibility of the learned $\blending$ can be advantageous with deep features.
The learned $\blending$ performs well when meta-learned but it has difficulty to accurately learn the angular dependencies in only 10 minutes of training otherwise.
\begin{wrapfigure}[26]{l}{0.43\textwidth}
	\vspace{-.2cm}
	\includegraphics[width=.4\textwidth]{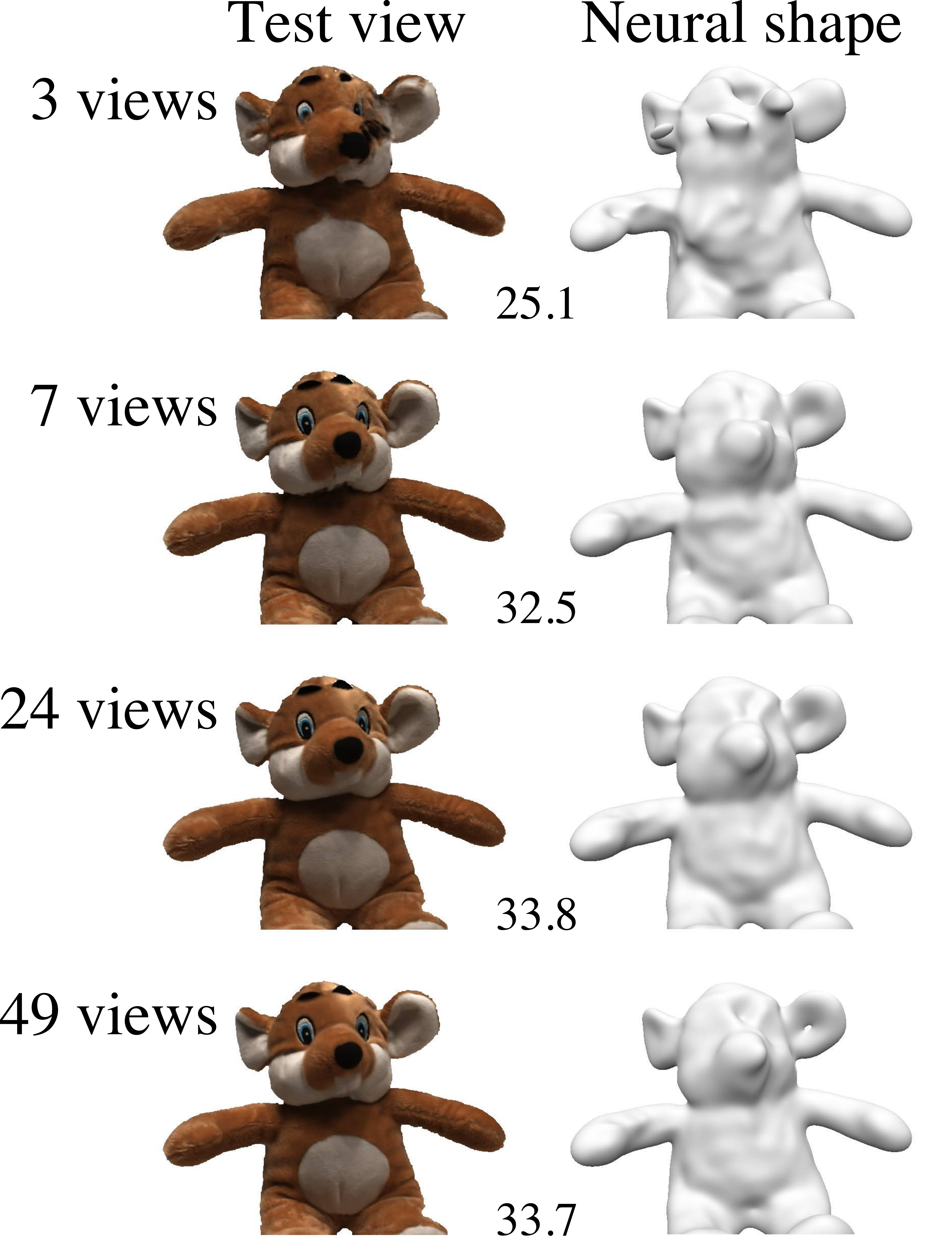} 
	\caption{\ours{} is robust to the number of views captured, which is essential in many applications where capturing a dense dataset is infeasible. In all cases, the learned $\shape$ provides an adequate support for projection of our encoded appearance features.}
	\label{fig:ablation}
\end{wrapfigure}
More details of this experiment and additional results are available in the supplementary document.

\paragraph{Input size ablation.}
We illustrate the robustness of \ours{} to low numbers of available training views in Figure ~\ref{fig:ablation}.
Here we see that our view synthesis quality decreases very gracefully with decreasing number of input views and it produces meaningful results even in the extreme case of three views. 
This is unlike COLMAP, which produces geometry with significant holes in occluded regions.
These holes are responsible for the highly variable performance of SVS* shown in Figure~\ref{fig:trade_off}.
Additionally, while SVS* is able to quantitatively perform well when evaluated within the ground-truth mask, the holes in the geometry result in inaccurate rendering masks and thus severely limit novel view synthesis in practice. Additional results showing this phenomenon are included in the supplementary document.
Since this ablation is performed on the DTU dataset, the input images capture only one side of the object. However, our method is applicable in scenarios where images are distributed $360^\circ$ around the object. Additional results demonstrating this capability on the ShapeNet~\cite{shapenet2015} dataset are included in the supplementary document.
Our learned shape models are overly smooth relative to other methods and thus not particularly quantitatively accurate, but provide sufficient accuracy for the image-based feature blending method to model appearance information observed in synthesized views. The smoothness of the learned shape is dependent on the capacity of the CNN architecture used for the feature encoder and decoder and coordinate-based network modeling the surface, as shown in the supplementary document.
We opt to use higher capacity feature encoder and decoders in order to model sharp image details for novel view synthesis, which is responsible for the trade-off of quantitative accuracy of the neural shape representation.

\paragraph{Additional results.}
To show that \ours{} is robust to datasets beyond DTU, we evaluate using the multi-view dataset in NLR~\cite{kellnhofer2021neural}. This dataset has been publicly released, and while the faces of the subjects in the dataset are personally identifiable, the subjects are the authors of NLR and have provided their consent of making this dataset public.
The scenes in this dataset each have 22 multi-view images, taken with various cameras. We opt to use the final 6 views taken with high-resolution cameras to train our representation.
We use 5 scenes in this dataset to learn the meta initialization $\params_0^*$, and test on a withheld sixth scene.
The meta learning in this case leads to significantly improved performance in both PSNR and perceptual quality, despite only being able to learn a prior from a small number of scenes.
This trend demonstrates that when the meta-training data more accurately covers the testing domain, the prior learned through meta learning is more effective at speeding up training time. This trend is further demonstrated in the supplementary document on the ShapeNet dataset, where the meta-training dataset is significantly larger and more closely related to the specialization data than in the case of DTU. With more comprehensive multi-view training datasets, this trend shows that \ours{} could even further speed up neural representation training time.

\begin{figure}
	\includegraphics[width=\textwidth]{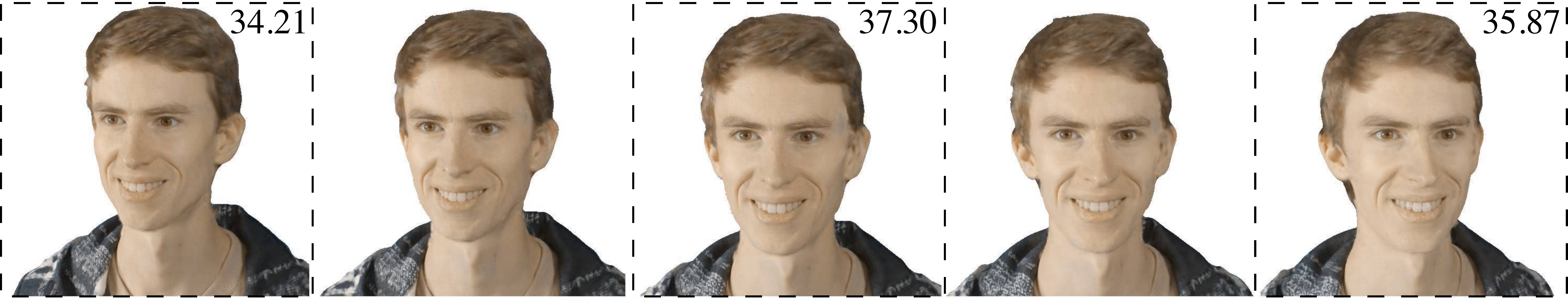} 
	\caption{We compute additional results using the NLR dataset. The frames highlighted are supervised, and the intermediate frames are interpolated viewpoints for this \ours{} model. Please see the supplement for additional results.}
	\label{fig:additional_results}
\end{figure}

In Table~\ref{tab:quantitative_data2}, we report our fit on the training frames after 30 minutes of training as benchmarked on our system.
Additionally, we show interpolated frame results in Figure~\ref{fig:additional_results}, demonstrating that \ours{} is capable of generalizing to this scene and producing convincing novel-view synthesis results. Additional implementation information and results are included in the supplementary document.

\begin{table}
	\centering
	\begin{tabular}{lcc}
		& 30 min. PSNR (LPIPS) & Convergence PSNR (LPIPS) \\
		\midrule
		\oursnm{} & 28.22 (0.145) & 35.54 (0.046) \\
		\ours{} & \textbf{31.12 (0.083)} & \textbf{37.55 (0.034)} \\
		\bottomrule
	\end{tabular}
	\vspace{0.5em}
	\caption{Table comparing PSNR score on the supervised views of the NLR dataset. Since there are only 6 training views, we use all for training and report PSNR averaged on the three views shown in Figure~\ref{fig:additional_results}, which shows that we are able to interpolate between these views.}
	\label{tab:quantitative_data2}
\end{table}

\section{Discussion and Conclusion}
\label{sec:discussion}
The fundamental problem of learning 3D scene representations from sparse sets of 2D images is rooted in machine learning, computer vision, and computer graphics. We provide an answer to this problem by proposing a novel parameterization of a 3D scene, and an efficient method for inferring these parameters from observations using meta learning. We demonstrate that our representation and training method are able to reduce representation training time consistently and render at real-time rates without sacrificing on image synthesis quality. This opens several exciting directions for future work in efficient training and rendering of representations, including using more advanced generalization methods to learn representations in real time. With this work, we make important contributions to the field of neural rendering.

\paragraph{Limitations and future work.}
While our method is able to produce compelling novel view synthesis results in a fraction of the time of other methods, we note that there are a few shortcomings. 
Specifically, in order to bootstrap the learning of a neural shape quickly, object masks are required to supervise the ray-mask loss. While these can be computed automatically for some data, this poses a challenge in cluttered scenes, or applications which could generalize to arbitrary scenes.
Additionally, all of our experiments have used known camera poses to reconstruct the shape. Future work on jointly optimizing camera pose with our representation is certainly possible, and a step in the direction for general view synthesis.
Our method is also limited by memory consumption, since the CNN feature encoder/decoders process the entire image at a time. This method could likely be improved by shifting to training and evaluating on image patches for higher resolution rendering.
Finally, our method tends to produce overly-smoothed shape models, which, while beneficial for feature aggregation, are not always representative of high-frequency scene geometry. This highlights one fundamental trade-off: the capacity of the feature generation method versus the quality of the shape. With feature or color generation methods which are sufficiently regularized~\cite{yariv2020multiview,kellnhofer2021neural}, the model has no choice but to explain observed details in the neural shape model. We opt to utilize the full capacity of the CNN feature processing, as learning a detailed neural shape is slower than modeling fine details with features.

\paragraph{Conclusion.}
The question of how to trade-off image synthesis quality with representation training and rendering time is of paramount importance to engineers, producers, or any other users of neural rendering technology. In the space of neural rendering methods, this work takes steps towards making representation learning and rendering more practical by optimizing this trade-off. Our novel scene parameterization and generalization method may provide a starting point for future work in optimizing this trade-off: speeding up representation training and rendering time and bringing modern neural rendering to the forefront of industry standard techniques.

\section{Broader Impact}
\label{sec:bi}

Methods such as \ours{} for learning 3D scene representations from 2D representations allow for photorealistic image synthesis using only collections of other images. We have shown that \ours{} improves upon the axes of training and rendering time of these representations, and as such may make it less computationally restrictive to use for individuals  who want to learn and use 3D models from only collections of easily acquirable images. While this proliferation of neural rendering technology may be extremely helpful for many, it has the potential for misuse. As with any synthesis method, the technology could enable approaches to synthesis of deliberately misleading or offensive images, posing challenges similar to those posed by generative-adversarial models.

\section*{Acknowledgments and Disclosure of Funding}
Alexander W. Bergman was supported by a Stanford Graduate Fellowship. Gordon Wetzstein was supported by a PECASE by the ARO. Other funding was provided by the NSF (1839974).

{\small
	\bibliographystyle{unsrtnat}
	\bibliography{references}
}



# 1 Additional Implementation Details

As described in the main text, we plan to release all code used to obtain the results for our method. All data used has been made publicly available by their authors. We use PyTorch for all implementation, and evaluate all of our methods using our internal server consisting of four Nvidia Quadro RTX8000 GPUs and six Nvidia Quadro RTX6000 GPUs, which we used a subset of. Due to our limited resources and requirement of training our method and many baselines to convergence, we opt to report error bars with respect to multiple different testing scenes instead of different random seeds. Each of these evaluations is run with a randomly generated seed. Implementation details on network architectures and hyperparameters for the DTU [1, 2] and NLR [3] datasets are included in Sections 1.1 and 1.2 respectively.

## 1.1 DTU dataset

**Data.** For each DTU scene, we use 7 of the ground truth 49 images for training. The view IDs of each of these images selected from the DTU dataset are: $[1, 9, 17, 25, 33, 41, 47]$. These views roughly image all parts of the object, but are not dense. The views held out for testing are views $[12, 32, 40]$. The images and ground truth masks are downsampled to resolution $800 \times 600$ as in [4], and all training and evaluation is performed on this resolution.

**Network architectures.** To represent our neural shape $\Phi_\theta$, we use a 5-layer MLP with 128 hidden units per layer. Multiple architecture widths and depths were considered before finding this architecture which maximized the trade-off between evaluation speed and reconstruction quality for the DTU objects. The source image encoder $E_\xi$ is implemented as a ResNet [5], using the same architecture in SVS [4]. This network consists of a ResNet18 network with 4 residual blocks, each consisting of Conv2d-BatchNorm2d-ReLU-Conv2d-BatchNorm2d network layers, where the first Conv2d downsamples (or upsamples) the image resolution by half in each dimension. Each of the skip connection layers in the network consist of a single Conv2d-ReLU network layer pairing. The output number of features is set to $d = 16$. The target image decoder $D_\psi$ is implemented as a UNet [6], with 3 down/up-sampling layers. Each downsampling block consists of Conv2d-ReLU-Conv2d-ReLU-AvgPool2d network layers. The intermediate number of channels after each block is: $[64, 128, 256]$. The learned aggregation function is implemented as a 5-layer MLP with 32 hidden units per layer, which maps each feature in $\mathbb{R}^{16}$ and its target viewing direction in $\mathbb{R}^3$ to the aggregation weight in $\mathbb{R}$.

**Pre-training.** The encoder and decoder networks are pre-trained using the FlyingChairs2 [7, 8] dataset. This dataset consists of pairs of images and ground-truth optical flow. To pre-train these networks, we apply two losses using image pairs and optical flow. The first loss feeds the image through the encoder, and the output features through the decoder, and ensures that the encoder and decoder are approximate inverses of each other. The second loss takes in the input image, warps the features, and then decodes the warped features into an image which is supervised by the warped image. This loss ensures that the encoder/decoder pair actually learn features representative of the image, and not to find some way to simply pass the input image through the feature bottleneck. This pre-training on the encoder and decoder especially helps NLR++ and SVS*, but MetaNLR++ is able to learn a prior over features using only meta learning on the DTU scenes. The shape network is pre-trained using a procedural sphere of radius 1.

**Training parameters.** When optimizing a single MetaNLR++ or NLR++ model to represent a DTU scene, a learning rate of $\eta_1 = 1 \times 10^{-4}$ was used to train $\Phi_\theta$, and a learning rate of $\eta_2 = 5 \times 10^{-5}$ was used for $E_\xi, D_\psi, \Gamma_\zeta$. The increased learning rate for the encoding, aggregation, and decoding functions encourages the network to learn to model appearance with deep features rather than with geometry, which encourages faster convergence. Learning rate $\eta_1$ is decreased by half at iteration numbers: $[500, 1000, 3000, 7000, 15000, 31000]$. Learning rate $\eta_2$ is decreased by half every 2000 iterations, consistently.

Other training parameters relate to the loss function applied and determining if features are occluded. We use a $\mathcal{L}_M$ weight of $\lambda_1 = 1 \times 10^2/\alpha$, where $\alpha$ is the mask softness parameter. The value of $\alpha$ is set at 50, and is doubled at iteration numbers $[2000, 4000, 6000]$. This enforces the mask to be more and more binary as training goes on. We use a $\mathcal{L}_E$ weight of $\lambda_2 = 3.0$ for all DTU experiments, which does not decay. This is applied on randomly sampled points in the unit cube which our scene

representation $\Phi$ lies. To determine whether sphere-traced features are occluded, we check whether or not sphere-traced surface points from the target and source views have an L2-distance smaller than threshold $\tau$. We start with $\tau = 1 \times 10^{-3}$, which changes to $\tau = 1 \times 10^{-4}$ at 5000 iterations, and $\tau = 1 \times 10^{-5}$ at 10000 iterations. This encourages the occlusions to be more strict as the shape quality improves.

Additionally, as described in the main text, one key component which makes our method converge quickly is the ability to balance shape and feature network optimization. One way this is done is by not computing shape gradients on each iteration, and using previously cached surface points from each view to determine feature occlusions. For the first $t_1 = 50$ and every $t_2 = 7$ iterations thereafter, both shape and feature encoder/decoders are optimized as described in the image formation model. On these iterations, sphere-traced surface point locations are cached for each view. On iterations where the shape is not trained, these sphere-traced surface point locations are used to determine occlusions, and create a target image which can be used to update $E_\xi, D_\psi, \Gamma_\zeta$. All views are sphere-traced at initialization, in order to serve as the initial cached surface values. Although these points may not be completely accurate, there is benefit in optimizing the feature processing networks more often as the shape evolves, as they can better learn to blend features for this specific scene. When optimizing both shape $\Phi_\theta$ and feature processing networks $E_\xi, D_\psi, \Gamma_\zeta$, we use a batch size of all 7 input images. However, gradients for the shape are only computed for 4 of these images – the remaining 3 images use cached surface point locations, as previously described.

Other relevant implementation details are: occluded features are set to value zero, and thus contribute nothing to the weighted sum feature aggregation. The batch gradients, as previously described, are computed by summing loss terms for each of the 4 rendered target images. Finally, all optimization uses the Adam [9], and we have found that using other optimizers significantly decreases the performance. We expect that this is because we have not spent a significant amount of time tuning every hyperparameter (due to the large number of parameters), and the Adam optimizer is robust to some of these choices. We expect that with further hyperparameter tuning, our method could likely receive better results.

**Meta learning.** We train the meta-initialization using 15 training DTU scenes, distinct from the scenes which we evaluate using. The initialization is learned using the Reptile algorithm [10], which simply updates the initialization in the direction of the optimized weights for $m = 64$ steps of fitting one of the training objects. This inner loop optimization is also done using the Adam optimizer, with the same training parameters as the previous section for the first 64 steps. However, $t_1$ is set to 64, which allows for shape optimization on every step of the meta-learning. The meta-learning rate is set to $\beta = 1 \times 10^{-1}$.

**Ablations.** For the ablation study on the meta learning, we use the same parameters as normal training for each method, including for training the meta-learned initializations. When training the meta-learned initialization for only the shape, we use the same method as meta-learning all parameters, but only update the shape network weights.

For the ablation study on number of input views, we also use the same parameters as normal training for each method. For 3 views, we use views $[1, 25, 47]$. For 24 views, we use views $[1, 5, 9, 13, 17, 21, 25, 29, 33, 37, 41, 43, 47]$. For 49 views, we use all views $1 - 49$ but still withhold views $[12, 32, 40]$ (leading to only 46 training views). The PSNR is computed on withheld views $[12, 32, 40]$.

## 1.2 NLR dataset

**Data.** For the NLR scenes, we use the last 6 ground truth images to train our representation. The view IDs of each of these images is: $[16, 17, 18, 19, 20, 21]$. This is done because these images are taken from the same camera, and using images from different cameras with a encoder and decoder which learn image priors may result in undesirable artifacts. These images all consist of human faces, an important application area for 3D representation learning. As in NLR, we do not withhold any specific views for testing, and instead qualitatively evaluate the interpolated view results. The images and ground truth masks are downsampled to resolution $800 \times 600$ as for the DTU dataset, and all training and evaluation is performed on this resolution.

3

**Network architectures.** We find that the default network architectures proposed for DTU lead to representations which build too much of the image appearance into the feature processing networks, and too little into the shape. This results in artifacts around the nose of subjects, where parts of the face should be occluded (see Figure 3). Thus, we propose to encourage the network to represent more of the high-frequency details using the geometry built into $\Phi$ instead of the feature processing $E, D, \Gamma$.

To represent our neural shape $\Phi_\theta$, we use a 5-layer MLP with 256 hidden units per layer. The source image encoder $E_\xi$ uses the same architecture as the DTU case. The target image decoder $D_\psi$ also uses a UNet to decode the features, but this UNet consists of 2 down/up-sampling layers with intermediate channel sizes after each block as $[32, 64]$. This smaller size prevents giving the decoder too much capacity to inpaint missing details, and overfit to the training images without refining the shape model. The learned aggregation function uses the same architecture as the DTU case.

**Pre-training.** Unlike in the DTU case, we do not pre-train the encoder and decoder networks using the FlyingChairs2 dataset. However, the shape network is still pre-trained using a procedural sphere of radius 1.

**Training parameters.** Most training parameters from the DTU case are re-used for the NLR case. One change regarding the batch size: since there are only 6 images in the dataset, the total image batch size is 6. However, the batch size of 4 for shape gradient computation remains the same. The other change is regarding the shape training versus iterations trade-off. Since these shapes require more fine detail to result in high-quality novel view synthesis, we opt to optimize the shape more often. Thus, the parameters $t_1, t_2$ are adjusted to $t_1 = 100, t_2 = 3$. This, along with the neural shape network size, affects the fastest possible speed with which we can represent these objects.

**Meta learning.** The meta-initialization is trained using 5 NLR scenes, distinct from the scene which we evaluate using. The initialization is learned using the Reptile algorithm with the exact same parameters as the DTU case.

### 1.3 ShapeNet dataset

**Data.** Each ShapeNet scene consists of 24 views at resolution $64 \times 64$. We withhold three views with IDs $[7, 16, 23]$ for testing, and use the remaining views for training. We randomly select 624 chair objects from the official ShapeNet training split to serve as meta-training data for the chairs split, and select 604 car objects from the official ShapeNet training split to serve as meta-training data for the cars split. We select 3 random objects from the official ShapeNet test set for the chairs and cars split respectively to serve as meta-testing samples.

**Network architectures.** To represent our neural shape $\Phi_\theta$, we use a 5-layer MLP with 128 hidden units per layer. This was selected to remain consistent with the experiments on the DTU dataset. Following this, the same architecture is used for the source image encoder $E_\xi$ and target image decoder $D_\psi$ and learned aggregation function as in the DTU experiments.

**Pre-training.** The ShapeNet experiments use the same pre-trained models as the DTU experiments - i.e. the encoder and decoder networks are pre-trained using the FlyingChairs2 dataset, and the shape network is pre-trained using a procedural sphere of radius 1.

**Training parameters.** Most training parameters from the DTU case are re-used for ShapeNet. Since the images are small and GPU memory is not a limitation, the batch size is increased to use all training 21 images when reconstructing each target image. At each iteration, the batch size is set to 10 for shape gradient computation (10 of the 21 images are treated as target images during each iteration). The parameters $t_1, t_2$ are adjusted to $t_1 = 2,000, t_2 = 5$. The large value of $t_1$ is acceptable due to the low resolution of the images, and thus the shape optimization through sphere tracing does not slow down each iteration of optimization by much.

**Meta learning.** The initialization is trained using the Reptile algorithm with the same parameters as the DTU case besides learning rate, which is decreased to $\beta = 2 \times 10^{-2}$. The meta-training and meta-testing splits are described in the section on data.

## 1.4 Timing

The timing for all methods was computed using an Nvidia RTX6000 GPU. For our method, we compute the timing by adding the time the forward pass takes, the time the loss computation takes, the time the backward pass takes, and the time that the optimization update takes. We compute distinct timing for iterations with and without shape optimization. For each of these types of iterations, we sampled 100 iteration times and averaged them to come to an iteration time. These values were then extrapolated to compute all timing results. The timing results for each of the methods at training time is:

    NeRF [11]: 0.40 sec/iteration.

    IBRNet [12]: 0.33 sec/iteration.

    NLR [3]: 2.21 sec/iteration.

    SVS* [4, 13, 14]: 1.38 sec/iteration after 96 second mesh computation time.

    IDR [1]: 0.2 sec/iteration.

    NLR++/MetaNLR++: 2.19 sec/iteration with no shape optimization, 7.69 sec/iteration with shape optimization.

Note that while MetaNLR++/NLR++ iterations appear to take significantly longer than those of other methods, these methods require significantly less iterations to converge. This is because, when compared to NLR or IDR, MetaNLR++ optimizes the representation for multiple entire images instead of randomly selected rays. The ray-batch size is significantly larger, resulting in longer iteration times.

The timing results for each of the methods at rendering time is computed assuming that the mesh and encoders can be pre-computed. Thus, this only requires learned aggregation and decoder evaluation, for our method and SVS*. For NeRF and IBRNet, the full forward pass must be ran on a full-resolution image. For IDR and NLR, this time is based on obtaining the pre-computed lumigraph, as described in NLR.

    NeRF: 32 sec/frame.

    IBRNet: 33.3 sec/frame.

    NLR: 0.025 sec/frame.

    SVS*: 0.031 sec/frame.

    IDR: 0.025 sec/frame.

    NLR++/MetaNLR++: 0.031 sec/frame.

## 1.5 Sphere tracing

We use the sphere tracing implementation published in [1]. This implementation uses bidirectional sphere tracing to find the intersection of the surface defined by $\Phi_\theta$ and a ray. We limit the sphere tracer to only 8 steps, and mark rays with SDF value within $5 \times 10^{-5}$ of 0 as converged. To find the minimum value along each ray necessary for $\mathcal{L}_M$, we densely sample 40 evenly spaced SDF values along a ray.

## 1.6 Result details

Here we describe additional noteworthy details of how the results in the main paper were computed.

**Figure 2.** The PSNR and LPIPS scores are computed on saved models throughout the training process after the convergence. These values smoothed values $\hat{v}_i$ were obtained using exponential moving average smoothing on the original measured values $v_i$, defined by:

$$\hat{v}_i = \begin{cases} v_i & i = 1 \\ \alpha v_i + (1-\alpha)\hat{v}_{i-i} & i > 1, \end{cases} \quad (1)$$

where we use smoothing parameter $\alpha = 0.8$ for all methods besides IDR, which uses $\alpha = 0.3$ due to the much higher variance between plot points. Our definition of the smoothing parameter $\alpha$ implies that smaller $\alpha$ results in more smoothing, and larger $\alpha$ results in less dependence on previous values and thus less smoothing.

5

We train all methods until convergence. In the figure, IDR converges prior to the 10,000 total seconds. Thus, we extrapolate these times using the average of the last 7 iteration results. This results in the straight line plot shown for the IDR bar in this figure.

**Table 1.** To find the time to reach specified dB PSNR for each model, we evaluate the PSNR for all saved models. We then use the lowest model iteration which reaches a specific metric, and compute the convergence time as a function of iterations. While it is computationally infeasible to evaluate all metrics for each model iteration, we save models as often as possible for each method. Here, we describe the model saving schedule, where notation $[n, m]$ means we have a model every $n$ iterations until total iteration $m$ is reached.

> NeRF: $[1, 10], [10, 100], [100, 1000], [1000, 10000], [10000, \text{end}]$
> IBRNet: $[1, 10], [10, 100], [100, 1000], [1000, 10000], [10000, \text{end}]$
> NLR: $[5, 100], [50, 1000], [200, \text{end}]$
> SVS*: $[5, 50], [25, 250], [250, \text{end}]$
> IDR: $[100, \text{end}]$
> NLR++/MetaNLR++: $[5, 50], [25, 250], [250, \text{end}]$

**PSNR/LPIPS Computations.** All PSNR computations are computed using the ground truth image masks and the method used in the code of SVS [4]. This method computes PSNR by element-wise multiplying (defined as $\circ$) ground truth and target image by the binary mask, and then computing PSNR on these images:

$$\text{PSNR}(\hat{I}_t, I_t, M_t) = 20 \log_{10}(1.0) - 10 \log_{10}((M_t \circ (\hat{I}_t - I_t))^2). \quad (2)$$

While this method biases the PSNR higher by using the masked values for error computation, it is consistently used for all baseline methods and thus comparisons are standardized. LPIPS metrics are also computed on the masked images which have also been multiplied by the ground truth mask, $\hat{I}_t \circ M_t$.

## 2 Baseline Implementation Details

**NeRF.** We use the NeRF implementation provided by the authors [11]. The training and evaluation were done with the same set of views and in the same $800 \times 600$ pixel resolution as for the scene fitting in our own method. Inspired by the provided *config_fern.txt* configuration, we evaluate 64 samples along each ray in the coarse phase and 128 samples in the fine phase and we process 2,048 rays in each batch. The training to convergence for each test scene was stopped after 200,000 solver steps as recommended by the authors.

**IBRNet.** We use the IBRNet implementation provided by the authors [12]. We use parameters from the provided *pretrain.txt* configuration to learn the generalized model for 250,000 solver steps on the same set of 15 training DTU scenes as for our own method. We then fine-tune the pre-trained model on the seven training views of a specific test scene for another 60,000 steps as defined in the *finetune_llff.txt* configuration. The training and evaluation is performed in the same $800 \times 600$ pixel resolution as for our own method.

**IDR.** We use the IDR implementation provided by the authors [1]. The training and evaluation is performed with the same training and test views and the same $800 \times 600$ pixel resolution as for our own method. We use the configuration for the DTU dataset provided by the authors which trains each scene for 2000 epoch where each epoch samples one batch of rays once for each of the seven training input views. This yields 14,000 solver steps for our input scenario.

**NLR.** We received access to the NLR implementation from the authors [3], and have used this for training comparison models. The training and evaluation is performed with the same training and test views and the same $800 \times 600$ pixel resolution as for our own method. The converged models were sampled after 100,000 solver steps, at which point loss curves stopped decreasing.

**SVS\*.** We use our own mesh-based method implementation inspired by SVS [4], as described in the text. We use COLMAP 3.6 [13, 14] to reconstruct surface mesh from the subset of 7 training views with known intrinsic parameters for each of our test DTU scenes [2]. The input images are resampled to the same $800 \times 600$ pixel resolution as for training of our own method. We follow the settings described in previous work [1–3]. We remove background points from the fused point-cloud using the object masks and we performed Poisson reconstruction of the surface with trim parameter set to 7. For training the encoder, decoder, and aggregation function $E_\xi, D_\psi, \Gamma_\zeta$, we use the same parameters and architecture as NLR++. However, since no shape optimization is done, every iteration optimizes the feature processing networks.

**MetaNLR.** We received access to the NLR implementation from the authors [3], and have wrapped this implementation with an outer loop which learns the initialization of the shape and color networks using Reptile [10]. The meta-training and testing are performed with the same meta-training and meta-test scenes, and the specialization to the test scenes is done using the same training and testing views (at $800 \times 600$ resolution) as in our method. We independently tune the hyperparameter for meta-learning rate, which we set as $5 \times 10^{-2}$, and use the same amount of inner loop steps per meta-iteration as our method.

## 3 Additional Results

In this section we provide additional results of our method, and other baseline methods for various scenes.

### 3.1 DTU dataset

**Additional image renderings & SVS\* results.** In our attached video and Figure 1, we show additional results comparing novel view synthesis results from various methods. In the video, we see that after a short amount of time, our method produces the highest quality novel view synthesis results. Note that as mentioned in the main text, the SVS\* method's ability to synthesize novel views is limited by the inaccurate rendering masks in the COLMAP-produced geometry. This results in large holes in the rendered images. This phenomenon is less observed when quantitatively comparing rendered images using the ground-truth masks, as in the main text, but when rendering views without ground truth object masks, either the full image or rendered mask must be used to display images.

Note that the results in the video for SVS\* also exhibit flickering artifacts between views, which are not apparent in the results produced by the original SVS implementation [4]. However, in the experiments presented, the number of training views is limited to only 7, and thus the feature processing networks $(E_\xi, D_\psi, \Gamma_\zeta)$ are more prone to overfitting to the training views. This overfitting results in less smooth blending between the training views. Additionally, the original SVS implementation uses a graph attention network instead of a MLP for on-surface feature aggregation, which may provide smoother feature interpolation.

We also include IBRNet and MetaNLR in these comparisons. IBRNet method is able to blend input images together and produce qualitatively good results, but cannot be rendered in near real-time. MetaNLR improves upon the training speed of NLR, but is still not able to match the visual quality of MetaNLR++ in the case of fast training, demonstrating that both our novel scene parameterization and the application of meta learning are necessary for fast training of high quality view synthesis methods.

**Ablation results & shape quality evaluation.** In Figure 2, we provide additional image results from the meta learning ablation study. We show qualitative results from meta-learned initializations specialized for 10 minutes of training. We show permutations of MetaNLR++ with RGB pixels instead of an encoder and decoder, and with a learned and fixed aggregation function. We see that the learned aggregation function introduces distortion in training when used with RGB pixel values directly, but leads to improved results when used with deep feature blending. The methods which use RGB pixel values directly are unable to inpaint and correct errors in the geometry, and thus lead to missing colored areas around the object boundary and falsely blended or occluded features when the shape is not correct.

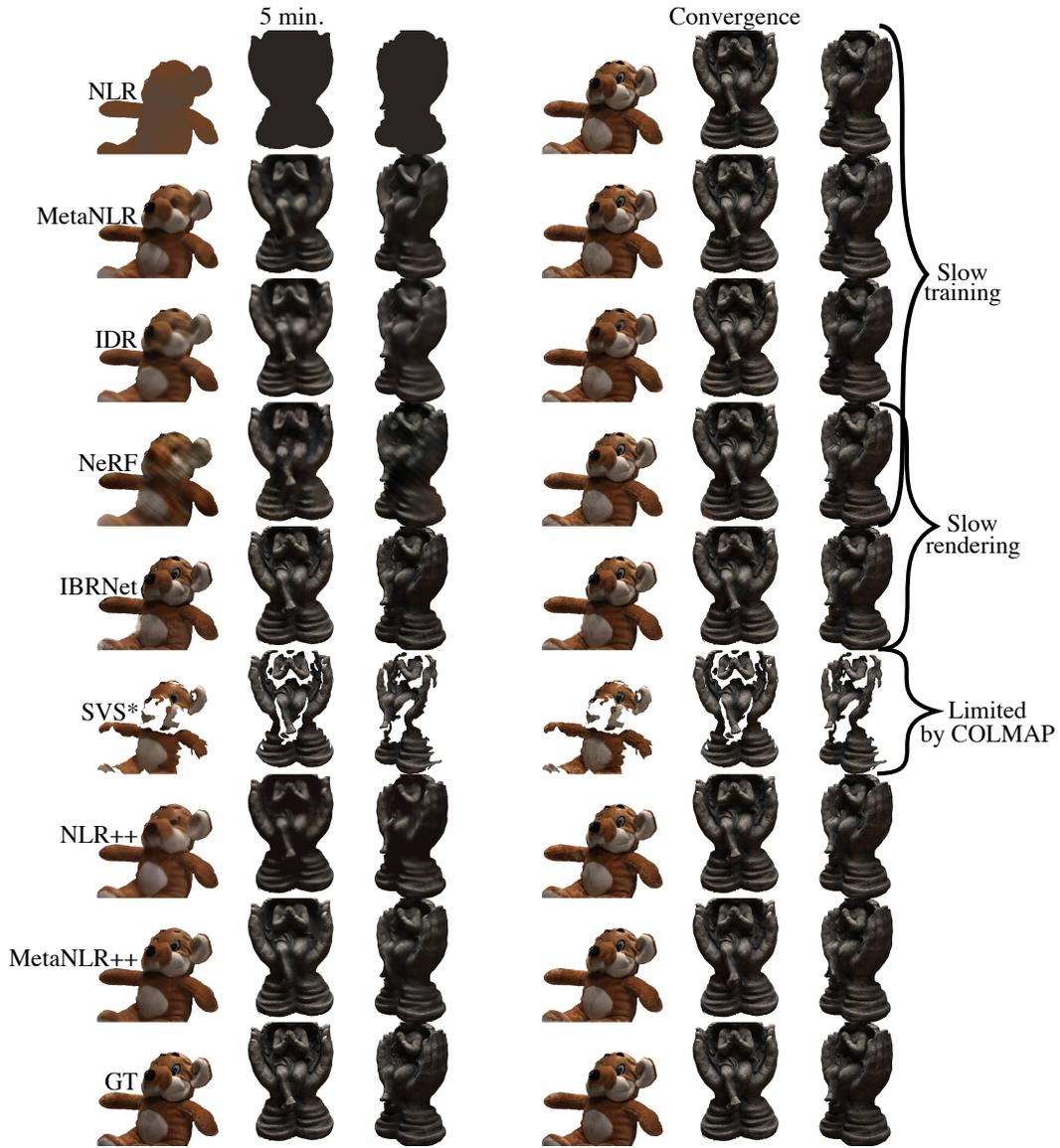

Figure 1: Additional qualitative results for all methods compared after 5 minutes of training and convergence. Here we apply the ground-truth background masks to all methods. Additionally, in the case of surface based methods, we also render test views using the masks extracted from the surface. This shows that the COLMAP results are limited by holes, especially in the mouse scene (DTU 105).

8

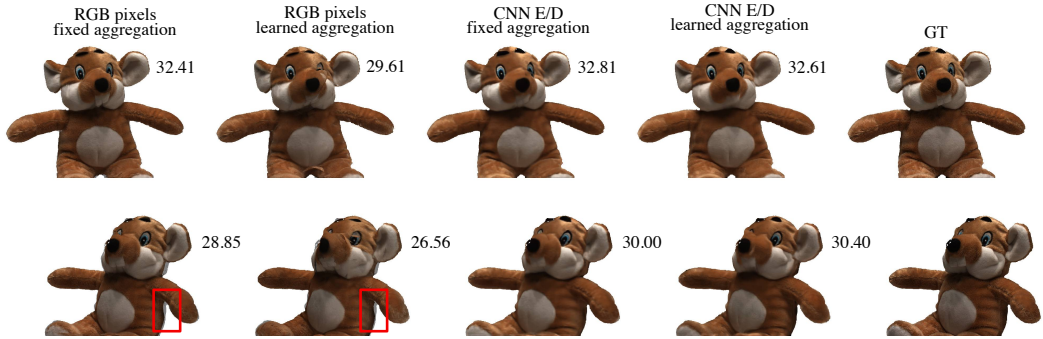

Figure 2: Comparison of the various ablated methods. In this comparison, we apply meta learning to all available networks, and show results after 10 minutes of training. The methods which use the RGB pixels produce sharp results, as the encoder and decoder do not need to be fine-tuned, but are not robust to errors in the geometry, as highlighted in the figure. These areas are inpainted in the version with a decoder.

|  | 5 min. Chamfer | 10 min. Chamfer | Convergence Chamfer |
| --- | --- | --- | --- |
| NeRF | 108.22* | 61.92 | *1.54* |
| IDR | **3.10** | **1.78** | **1.42** |
| NLR | 7.46 | 6.27 | 1.68 |
| COLMAP | 5.61 | 5.61 | 5.61 |
| NLR++ | 3.86 | 3.39 | 2.70 |
| MetaNLR++ | *3.26* | *2.78* | 2.28 |

Table 1: Table comparing Chamfer distances of the shapes produced from different methods at different times. These Chamfer results are averaged over the three DTU test scenes.
*For one of the DTU test scenes, a mesh was unable to be extracted, so this distance is the average of the other two scenes.

In Table 1, we compute the Chamfer distance for the methods shown in the main text. IDR performs well in reconstructing a high quality shape quickly, but as shown in the main text and Figure 1, rendered novel view synthesis is low-quality. This detail in shape is likely a result of the low capacity of the color representation, forcing the neural shape model to model differences in rendered views with geometry. MetaNLR++ achieves the second fastest shape convergence, significantly faster than volumetric methods such as NeRF, and faster than other neural surface based methods like NLR.

### 3.2 NLR dataset

For additional comparisons and convergence on the NLR dataset, please see our supplemental video.

**Failure cases.** As noted in the implementation details, when working with the NLR dataset, we increase the capacity of the neural shape representation and decrease the capacity of the feature decoder. Specifically, we increase the neural shape model $\Phi_\theta$ hidden dimension from 128 to 256 and decrease the number of UNet blocks in the decoder module $D_\psi$ from 3 to 2, with half the number of channels in each block. Without these architecture changes, the representation is unable to represent fine geometric details, and thus uses the decoder to decode the falsely blended features and inpaint the falsely occluded features to produce the training images. This results in overfitting to the training views. As shown in Figure 3, this is especially effective around the nose area of the face in the NLR dataset. Features here are blended when they should be occluded by the nose, resulting in blurriness. This blurriness is especially apparent in the interpolated views, which the decoder has not seen and thus cannot memorize the ground truth image.

### 3.3 ShapeNet dataset

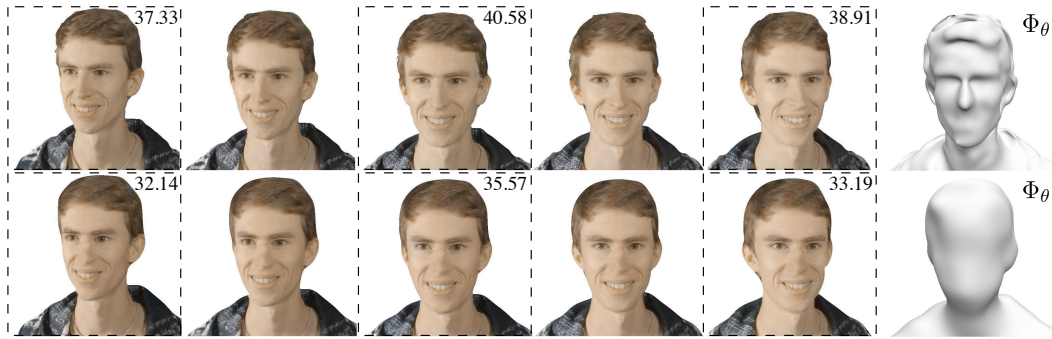

Figure 3: The first row shows the converged results on the NLR dataset with the architecture changes to $\Phi_\theta$ and $D_\psi$. The second row is a failure case, where the novel view synthesis quality interpolates poorly and is unable to fit the ground truth frames as well. This occurs because the shape model $\Phi_\theta$ does not have enough capacity to model the fine details of the face, and the $E_\xi, D_\psi, \Gamma_\zeta$ have too much capacity to overfit to training views.

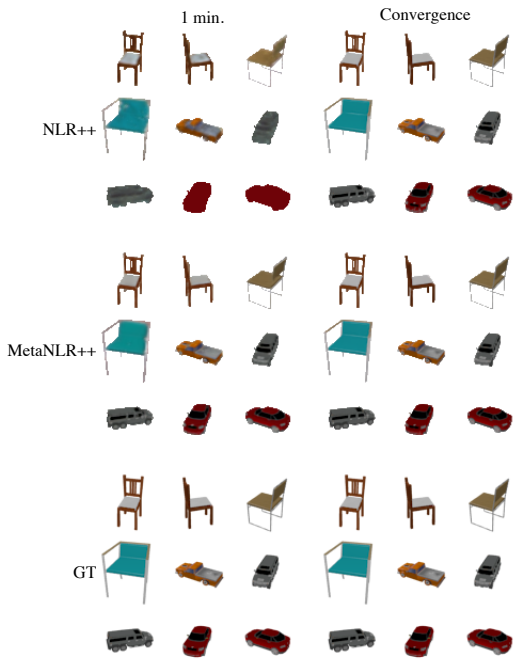

Figure 4: Qualitative examples from the cars and chairs splits of the ShapeNet dataset. These are shown after 1 minute of training and convergence for MetaNLR++ and NLR++.

We ablate the meta learning contribution by comparing MetaNLR++ to NLR++ on the cars and chairs splits of the ShapeNet dataset [15]. The relative performances are demonstrated in Table 2 by quantifying the time to reach a certain PSNR, and qualitative results are shown in Figure 4 after varying amounts of training. The testing and training views of the ShapeNet objects are distributed 360° around each object, demonstrating that MetaNLR++ is able to function well in this scenario.

From quantitative and qualitative results, we see that meta learning used in MetaNLR++ leads to a significant improvement over simply using the NLR++ scene parameterization. This continues the trend as described on the NLR dataset in the main paper, where a more comprehensive meta-training object dataset covering the testing object distribution leads to improved relative benefit of meta-learning. In the case of the cars split of the ShapeNet dataset, the lower variation between car objects than, for example, DTU objects leads to a stronger meta-prior learned and thus relatively faster training of MetaNLR++ compared to NLR++. This trend is observed to a lesser extent for the chairs split, where the objects are not as uniform as cars. We hypothesize that for future applications of MetaNLR++, larger meta-training datasets and specific applications could lead to even faster training.

10

| Cars Split | 30dB PSNR | 35dB PSNR | Maximum PSNR |
|---|---|---|---|
| NLR++ | 1.5 min. | 36.0 min. | 35.5dB |
| MetaNLR++ | **12.0 sec.** | **3.7 min.** | **37.9dB** |

| Chairs Split | 30dB PSNR | 35dB PSNR | Maximum PSNR |
|---|---|---|---|
| NLR++ | 3.6 min. | 125.0 min. | 37.0dB |
| MetaNLR++ | **1.5 min.** | **50.0 min.** | **40.7dB** |

Table 2: Comparison of the time to reach a specified PSNR level on the ShapeNet dataset for MetaNLR++ and NLR++.



\section*{Checklist}


\begin{enumerate}

\item For all authors...
\begin{enumerate}
  \item Do the main claims made in the abstract and introduction accurately reflect the paper's contributions and scope?
    \answerYes{}
  \item Did you describe the limitations of your work?
    \answerYes{See Section~\ref{sec:discussion}.}
  \item Did you discuss any potential negative societal impacts of your work?
    \answerYes{See Section~\ref{sec:bi}.}
  \item Have you read the ethics review guidelines and ensured that your paper conforms to them?
    \answerYes{}
\end{enumerate}

\item If you are including theoretical results...
\begin{enumerate}
  \item Did you state the full set of assumptions of all theoretical results?
    \answerNA{}
	\item Did you include complete proofs of all theoretical results?
    \answerNA{}
\end{enumerate}

\item If you ran experiments...
\begin{enumerate}
  \item Did you include the code, data, and instructions needed to reproduce the main experimental results (either in the supplemental material or as a URL)?
    \answerNo{We plan to release the code completely but have not yet with submission.}
  \item Did you specify all the training details (e.g., data splits, hyperparameters, how they were chosen)?
    \answerYes{See supplementary document and Section~\ref{sec:implementation_details}.}
	\item Did you report error bars (e.g., with respect to the random seed after running experiments multiple times)?
    \answerNo{As models take a significant amount of time to train, averaging over many seeds would be computationally impractical. We report standard deviation error bars with respect to multiple testing scenarios, which demonstrate the robustness of our method.}
	\item Did you include the total amount of compute and the type of resources used (e.g., type of GPUs, internal cluster, or cloud provider)?
    \answerYes{See Section~\ref{sec:implementation_details}.}
\end{enumerate}

\item If you are using existing assets (e.g., code, data, models) or curating/releasing new assets...
\begin{enumerate}
  \item If your work uses existing assets, did you cite the creators?
    \answerYes{}
  \item Did you mention the license of the assets?
    \answerNo{We were unable to find the license of these datasets, but the authors have explicitly published their datasets to the public. We also discuss the nature of the datasets including their content and consent in Section~\ref{sec:experiments}}
  \item Did you include any new assets either in the supplemental material or as a URL?
    \answerNo{}
  \item Did you discuss whether and how consent was obtained from people whose data you're using/curating?
    \answerYes{See Section~\ref{sec:experiments}}
  \item Did you discuss whether the data you are using/curating contains personally identifiable information or offensive content?
    \answerYes{See Section~\ref{sec:experiments}, and supplementary document for additional discussion on the datasets.}
\end{enumerate}

\item If you used crowdsourcing or conducted research with human subjects...
\begin{enumerate}
  \item Did you include the full text of instructions given to participants and screenshots, if applicable?
    \answerNA{}
  \item Did you describe any potential participant risks, with links to Institutional Review Board (IRB) approvals, if applicable?
    \answerNA{}
  \item Did you include the estimated hourly wage paid to participants and the total amount spent on participant compensation?
    \answerNA{}
\end{enumerate}

\end{enumerate}


\end{document}